\documentclass[journal]{IEEEtai}

\usepackage[colorlinks = true,
bookmarks = false,
linkcolor=red,
citecolor=blue,
urlcolor=blue,
anchorcolor = blue]{hyperref}

\usepackage{color,array}

\usepackage{cellspace}
\usepackage{cite}
\usepackage{amsmath,amssymb,amsfonts,nccmath}

\usepackage{algorithmic}
\usepackage{graphicx,balance}
\usepackage{subfig}
\usepackage{balance}
\usepackage{textcomp}
\usepackage{xcolor}
\usepackage{color}
\usepackage{url}
\usepackage{xspace}
\usepackage{booktabs}
\usepackage{comment}
\usepackage{acronym}
\usepackage[none]{hyphenat}
\usepackage{ragged2e}
\usepackage{tabularx}
\usepackage{tablefootnote}
\newcommand{\tabitem}{~~\llap{\textbullet}~~}

\usepackage{colortbl} 
 \usepackage[normalem]{ulem}
 \useunder{\uline}{\ul}{}

\newcommand{\etc}{etc.\@\xspace}

\acrodef{AI}{Artificial Intelligence}
\acrodef{XAI}{Explainable AI}
\acrodef{FL}{Federated Learning}
\acrodef{ML}{Machine Learning}
\acrodef{DL}{Deep Learning}
\acrodef{NLP}{Natural Language Processing}
\acrodef{CNNs}{Convolutional Neural Networks}
\acrodef{RL}{Reinforcement Learning}
\acrodef{RNNs}{Recurrent Neural Networks}
\acrodef{LAWS}{Lethal Autonomous Weapon Systems}
\acrodef{NLAWS}{Non-Lethal Autonomous Weapon Systems}
\acrodef{LOAC}{Laws of Armed Conflict}
\acrodef{NSLM}{Neuro-Symbolic Language Model}
\acrodef{RNKN}{Recursive Neural Knowledge Network}
\acrodef{DARPA}{Defense Advanced Research Projects Agency}
\acrodef{ANSR}{Assured Neuro Symbolic Learning and Reasoning}
\acrodef{NATO}{North Atlantic Treaty Organization}
\acrodef{DG}{Deep Green}
\acrodef{OODA}{Observe, Orient, Decide, Act}
\acrodef{RAID}{Real-time Adversarial Intelligence and Decision-making}
\acrodef{MAS}{Multi-Agent Systems}
\acrodef{LLMs}{Large Language Models}
\acrodef{RAG}{Retrieval Augmented Generation}

\renewcommand\IEEEkeywordsname{Keywords}
\begin{document}

\title{Neuro-Symbolic AI for Military Applications}

\author{Desta Haileselassie Hagos, \IEEEmembership{Member, IEEE}, and Danda B. Rawat, \IEEEmembership{Senior Member, IEEE} 
\thanks{This work was supported by the United States DoD Center of Excellence in AI/ML at Howard University under Contract number W911NF-20-2-0277 with the U.S. Army Research Laboratory (ARL). However, any opinions, findings, conclusions, or recommendations expressed in this document are those of the authors and should not be interpreted as representing the official policies, either expressed or implied, of the funding agencies.}
\thanks{D. H. Hagos and D. B. Rawat are with the DoD Center of Excellence in Artificial Intelligence and Machine Learning (CoE-AIML), College of Engineering and Architecture (CEA), Department of Electrical Engineering and Computer Science, Howard University, Washington DC, USA (e-mail: desta.hagos@howard.edu; danda.rawat@howard.edu).}

}

\markboth{Accepted at IEEE Transactions on Artificial Intelligence (TAI)}
{Hagos \MakeLowercase{\textit{et al.}}: Neuro-Symbolic AI for Military Applications}

\maketitle

\begin{abstract}

\ac{AI} plays a significant role in enhancing the capabilities of defense systems, revolutionizing strategic decision-making, and shaping the future landscape of military operations. Neuro-Symbolic \ac{AI} is an emerging approach that leverages and augments the strengths of neural networks and symbolic reasoning. These systems have the potential to be more impactful and flexible than traditional \ac{AI} systems, making them well-suited for military applications. This paper comprehensively explores the diverse dimensions and capabilities of Neuro-Symbolic \ac{AI}, aiming to shed light on its potential applications in military contexts. We investigate its capacity to improve decision-making, automate complex intelligence analysis, and strengthen autonomous systems. We further explore its potential to solve complex tasks in various domains, in addition to its applications in military contexts. Through this exploration, we address ethical, strategic, and technical considerations crucial to the development and deployment of Neuro-Symbolic \ac{AI} in military and civilian applications. Contributing to the growing body of research, this study represents a comprehensive exploration of the extensive possibilities offered by Neuro-Symbolic \ac{AI}. 

\end{abstract}

\begin{IEEEImpStatement}

The AI-powered battlefield of the future will be driven by Neuro-Symbolic \ac{AI}, revolutionizing warfare. Leveraging \ac{AI} in military decision-making processes enhances battlefield effectiveness and improves the quality of critical operational decisions. The combination of neural networks and symbolic reasoning has the potential to revolutionize military operations by significantly improving threat detection accuracy and enabling faster, more precise tactical decision-making. This paper provides a thorough analysis that offers valuable insights for researchers, practitioners, and military policymakers who are concerned about the future of \ac{AI} in warfare. Through a critical examination of existing research, key challenges are identified, and promising directions for future development are outlined. This aims to further empower the responsible deployment of Neuro-Symbolic \ac{AI} in areas such as optimized logistics, enhanced situational awareness, and dynamic decision-making. Furthermore, the advancements in Neuro-Symbolic \ac{AI} for military applications hold significant potential for broader applications in civilian domains, such as healthcare, finance, and transportation. This approach offers increased adaptability, interpretability, and reasoning under uncertainty, revolutionizing traditional methods and pushing the boundaries of both military and civilian effectiveness.

\end{IEEEImpStatement}


\begin{table}[!t]
\centering
\caption*{List of acronyms used in this paper.}
\def\arraystretch{1.13}
\begin{tabular}{l|l}
\toprule
\textbf{Acronym} & \hspace*{\fill} \textbf{Definition} \hspace*{\fill} \\ \midrule 
AI & Artificial Intelligence \\
ANSR & Assured Neuro Symbolic Learning and Reasoning \\
CNNs & Convolutional Neural Networks \\
DARPA & Defense Advanced Research Projects Agency \\
DG & Deep Green \\
DL & Deep Learning \\
FL & Federated Learning \\
LAWS & Lethal Autonomous Weapon Systems \\
LLMs &  Large Language Models \\
LOAC & Laws of Armed Conflict \\
ML & Machine Learning \\
MAS & Multi-Agent Systems \\
NATO & North Atlantic Treaty Organization \\
NLP & Natural Language Processing \\
NLAWS & Non-Lethal Autonomous Weapon Systems \\
NSLM & Neuro-Symbolic Language Model \\
OODA & Observe, Orient, Decide, Act \\
RAG & Retrieval Augmented Generation \\
RAID & Real-time Adversarial Intelligence and Decision-making \\
RL & Reinforcement Learning \\
RNKN & Recursive Neural Knowledge Network \\
RNNs & Recurrent Neural Networks \\
T\&E & Testing and Evaluations \\
V\&V & Verification and Validation \\
XAI & Explainable AI \\

\bottomrule

\end{tabular}

\end{table}

\section{Introduction}
\label{introduction}

\IEEEPARstart{T}{he} rise of \ac{AI} is rapidly reshaping the world, particularly in the realm of military and national security~\cite{sayler2019artificial, hoadley2018artificial}. \ac{AI} is already being used across a variety of military applications, ranging from intelligence gathering and surveillance to autonomous weapons systems, fundamentally altering military operations~\cite{feldstein2019global}. In the past, human cognitive and decision-making processes were crucial in military strategies. However, the integration of \ac{AI} has introduced unprecedented levels of sophistication and efficiency. One significant example of this transformation is the use of autonomous drones~\cite{cullen2011mq}. These drones can perform reconnaissance, surveillance, and even precision strikes with minimal human involvement~\cite{cullen2011mq}. \ac{AI} has also revolutionized cybersecurity. Neural networks and \ac{ML} algorithms are employed to detect and respond to cyber threats in real-time~\cite{wirkuttis2017artificial}. This is critical for the defense community, as cyberattacks have become increasingly prevalent in the digital age~\cite{wirkuttis2017artificial}. Beyond traditional battlefields and cyberspace, \ac{AI} is also playing a pivotal role in enhancing logistical and strategic planning~\cite{plan2016national}. \ac{AI} can improve the efficiency and effectiveness of military logistics and supply chain management systems, ensuring that supplies are delivered to the right place at the right time~\cite{tsadikovich2010ai}. \ac{AI} can also be used to develop realistic training and simulation environments for soldiers, helping them to develop the skills and knowledge they need to operate effectively in combat~\cite{campbell1997use, erickson1985fusing}. Another significant impact of \ac{AI} in military settings is the development of advanced intelligence gathering and surveillance capabilities~\cite{feldstein2019global, scharre2018army, hare2016future}. AI-powered systems can process vast amounts of data from a variety of sources, including satellites, drones, and social media, allowing militaries to track enemy movements, identify potential threats, and assess battlefield conditions in real-time~\cite{feldstein2019global, hare2016future}. \ac{AI} is also a driving force in the development of cyberwarfare capabilities~\cite{borah2015cyber}, which involve using computer networks to target an enemy's computer systems and infrastructure~\cite{borah2015cyber, ashraf2021defining}. \ac{AI} can be used to develop cyberweapons that are more sophisticated and effective than traditional cyberweapons~\cite{caplan2013cyber}. In the 1950s and 1960s, researchers developed two main approaches to \ac{AI}: \emph{symbolic AI} and \emph{connectionist AI}~\cite{dinsmore2014symbolic}.

\begin{figure}[!h]

 \includegraphics[width=\linewidth]{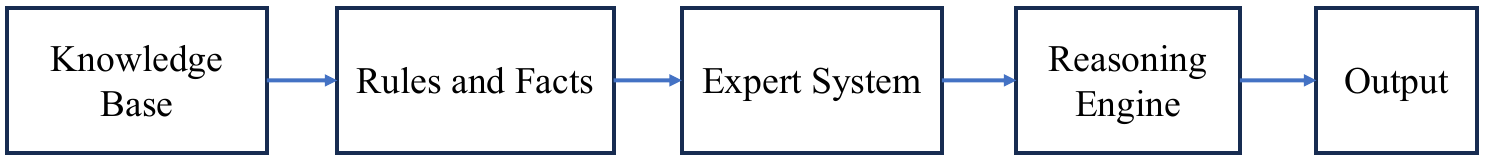}
 \caption{The \emph{Knowledge Base} contains rules and facts which are used to represent explicit knowledge. \emph{Rules and Facts} represent logical statements or heuristics that encode knowledge and guide reasoning. The \emph{Expert System} utilizes the knowledge base and reasoning engine to make decisions or provide solutions. The \emph{Reasoning Engine} involves the process of using logical inference to draw conclusions or perform tasks based on the rules and facts in the knowledge base. The final \emph{Output} represents the results or decisions made by the Symbolic \ac{AI} system based on its knowledge and rules.}

 \label{Fig:Symbolic_AI}
\end{figure}

\vspace{1.0ex}
\noindent \textbf{Symbolic AI}. Symbolic \ac{AI} is a traditional approach to \ac{AI} that focuses on representing and rule-based reasoning about knowledge using symbols such as words or abstract symbols, rules, and formal logic~\cite{goel2022looking, dinsmore2014symbolic, russell2010artificial, khorasani2008artificial}. Symbolic \ac{AI} systems rely on explicit, human-defined knowledge bases that contain facts, rules, and heuristics. These systems use formal logic to make deductions and inferences making it suitable for tasks involving explicit knowledge and logical reasoning. Such systems also use rule-based reasoning to manipulate symbols and draw conclusions. Symbolic \ac{AI} systems are often transparent and interpretable, meaning it is relatively easy to understand why a particular decision or inference was made. In symbolic \ac{AI}, knowledge is typically represented using symbols, such as words or abstract symbols, and relationships between symbols are encoded using rules or logical statements~\cite{dinsmore2014symbolic}. As shown in Figure~\ref{Fig:Symbolic_AI}, Symbolic \ac{AI} is depicted as a knowledge-based system that relies on a knowledge base containing rules and facts. It uses logical inference to perform reasoning and decision-making tasks.

\begin{figure}[!h]

 \includegraphics[width=\linewidth]{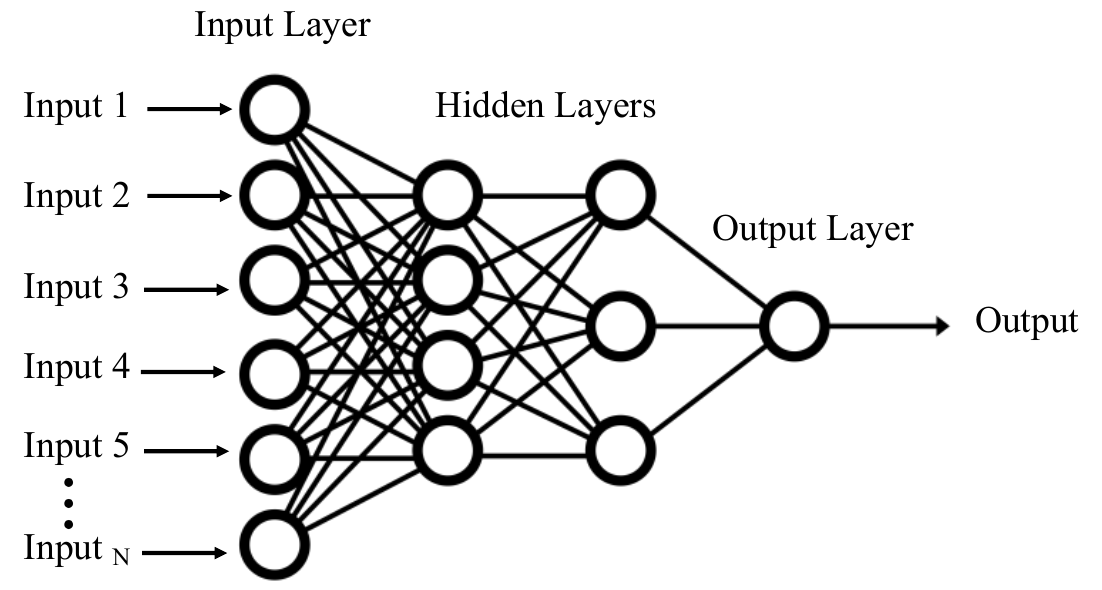}
 \caption{Data is fed into the network as input, and it processes the input data through multiple layers of neurons with adjustable weights. The network learns from data to produce output, making it well-suited for tasks involving pattern recognition and complex data processing.}
 \label{Fig:Connectionist_AI}
\end{figure}

\vspace{1.0ex}
\noindent \textbf{Connectionist \ac{AI}}. On the other hand, connectionist \ac{AI} systems leverage artificial neural networks and distributed representation to learn complex patterns from data and excel in tasks involving pattern recognition and complex data processing~\cite{goel2022looking, haykin1998neural, bishop2006pattern, lecun2015deep}. These systems, generally neural networks, are considered a black box due to their lack of interpretability. This is because it is challenging to interpret how they arrive at a particular decision. This lack of transparency is a trade-off for their ability to handle complex, high-dimensional data. The basic building block of a connectionist model is the artificial neuron, and these neurons are organized into layers. The most common type of connectionist model is the feedforward neural network which can be represented as shown in Equation~\ref{Eq:feedforward-connectionist-model} where $N_L$ is the number of neurons in layer $L$ and $f$ is the activation function. Here, $x_i$ is the input to the network,  $w_{i j}$ represents the weight connecting neuron $i$ in layer $L$ to neuron $j$ in layer $L+1$, $b_j$ denotes the bias term for neuron $j$ in layer $L+1$, and $a_j$ represents the activation of neuron $j$ in layer $L+1$. As shown in Figure~\ref{Fig:Connectionist_AI}, connectionist \ac{AI} is depicted as a neural network organized in layers of interconnected neurons. These networks learn from data by adjusting the strength of connections or weights between neurons. In a connectionist \ac{AI} setting, knowledge is distributed across the connections and weights of the neural network (see Figure~\ref{Fig:Connectionist_AI}). Information is processed in a parallel and distributed manner, allowing neural networks to capture complex patterns and relationships in data~\cite{hikosaka1999parallel, smolensky1987connectionist}. Connectionist \ac{AI} learns from large datasets through techniques like backpropagation, where errors are propagated backward through the network to adjust the weights~\cite{goodfellow2016deep}. It excels in tasks like image and speech recognition, \ac{NLP}, and \ac{RL}. \ac{CNNs} for image recognition~\cite{krizhevsky2012imagenet}, \ac{RNNs} for sequence modeling~\cite{lecun2015deep, goodfellow2016deep}, and Transformers for \ac{NLP}~\cite{wolf2020transformers} are examples of connectionist \ac{AI}.

\begin{equation}
a_j=f\left(\sum_{i=1}^{N_L} w_{i j} x_i+b_j\right)
\label{Eq:feedforward-connectionist-model}
\end{equation}

\begin{figure}[!h]

 \includegraphics[width=\linewidth]{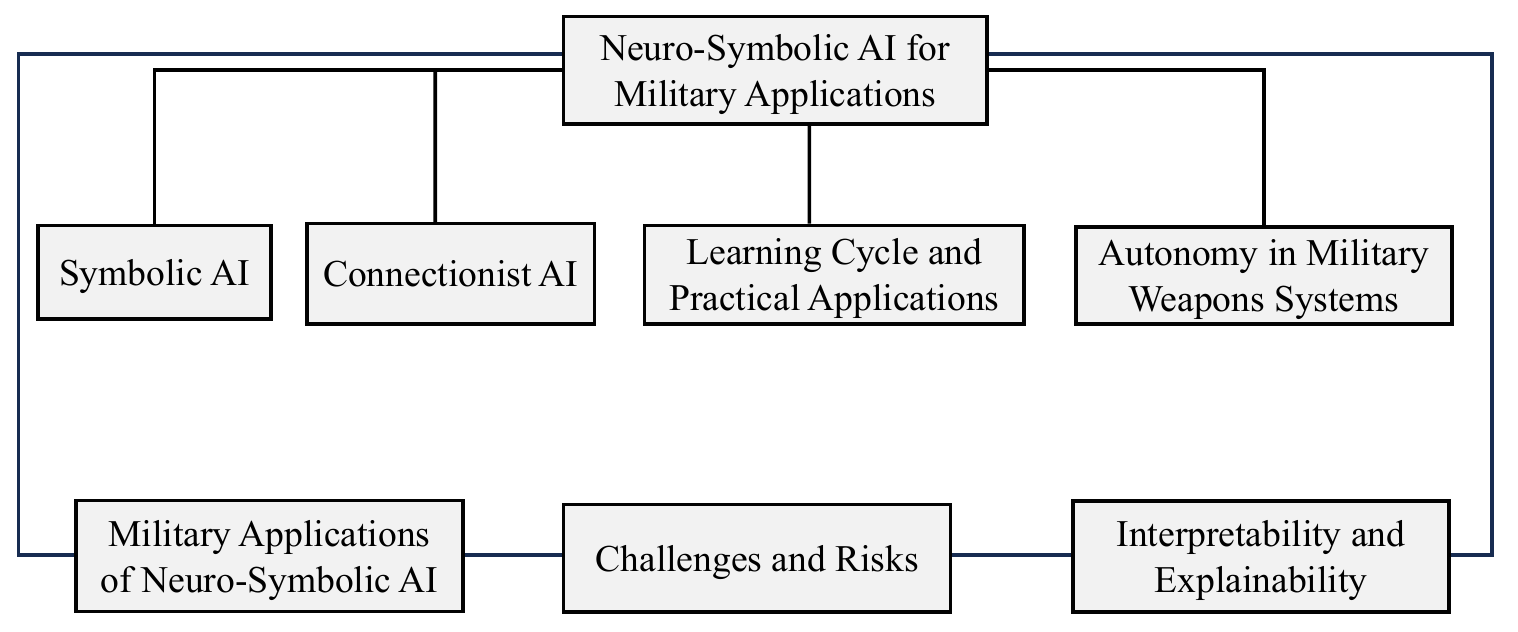}
 \caption{Taxonomy of the main topics covered in this paper. }
 \label{Fig:Taxonomy}
\end{figure}

This paper explores the potential applications of Neuro-Symbolic \ac{AI} in military contexts, highlighting its critical role in enhancing defense systems, strategic decision-making, and the overall landscape of military operations. Beyond the potential, it comprehensively investigates the dimensions and capabilities of Neuro-Symbolic \ac{AI}, focusing on its ability to improve tactical decision-making, automate intelligence analysis, and strengthen autonomous systems in a military setting. The paper addresses ethical, strategic, and technical considerations related to the development and deployment of Neuro-Symbolic \ac{AI} in the military. It identifies key challenges and proposes promising directions for future development, emphasizing responsible deployment in areas such as logistics optimization, situational awareness enhancement, and dynamic decision-making. Figure~\ref{Fig:Taxonomy} shows the taxonomy of the main topics covered in our work. In Table~\ref{table:comparison}, we present a comparison between our comprehensive exploration of Neuro-Symbolic \ac{AI} for military applications and existing research by highlighting key distinctions and contributions in the context of our work.

\begin{table*}

\caption{Comparison of our work to relevant existing works in the field.}
\label{table:comparison}
\begin{tabularx}{\linewidth}{r|l| X} 
\midrule
\textbf{Year} & \textbf{Publications} & \textbf{Main Research Focus and Scope}\\\midrule

2024 & Our Paper & \tabitem Represents a comprehensive exploration of the extensive possibilities of Neuro-Symbolic AI in military applications.

\tabitem Our work positions Neuro-Symbolic \ac{AI} as a game-changer for the future of military operations, promoting national security and operational effectiveness.

\tabitem Comprehensively addresses the ethical, strategic, and technical considerations crucial to the development and deployment of Neuro-Symbolic \ac{AI} in military settings.

\tabitem Outlines the strategic implications of integrating Neuro-Symbolic \ac{AI} into military operations under dynamic and complex situations. \\ \midrule

2023 & Ref~\cite{sheth2023neurosymbolic} & \tabitem Discusses the need for symbolic mappings to support explainability and control in cognitive functions, especially in safety-critical applications like healthcare and autonomous driving.\\ \midrule

2023 & ~\cite{gaur2024building} & \tabitem Emphasizes the importance of building AI systems that are consistent, reliable, explainable, and safe to ensure trust among users and stakeholders. The paper seeks to demonstrate how Neuro-Symbolic approaches can enhance trust and address challenges associated with Large Language Models (LLMs).\\ \midrule

2022 & Ref~\cite{yu2022probabilistic} & \tabitem Explores a model that combines probabilistic graphical modeling techniques with neural-symbolic reasoning to enhance visual relationship detection in computer vision applications. \\ \midrule

2022 & Ref~\cite{hitzler2022neuro} & \tabitem Provides an overview of the history, challenges, and various aspects of Neuro-Symbolic \ac{AI}, along with a list of key resources and literature pointers for further exploration. \\ \midrule

2022 & Ref~\cite{alvaroANSR} & \tabitem Focuses on developing hybrid AI algorithms that integrate symbolic reasoning with data-driven learning to create more trustworthy and dependable autonomous systems for military applications. \\ \midrule

2021 & Ref~\cite{bistron2021artificial} & \tabitem Provides a comprehensive analysis of AI in military systems and its impact on sense of security of citizens. \\ \midrule

2020 & Ref~\cite{amizadeh2020neuro} & \tabitem Tackles the problem of visual reasoning, specifically focusing on separating the perceptual visual aspects from the reasoning aspects in tasks like visual question answering. \\ \midrule

2018 & Ref~\cite{voogd2018neuro} & \tabitem Argues that automating situational understanding from diverse information sources significantly improves decision-making in complex environments. \\ \midrule

2018 & Ref~\cite{schubert2018artificial} & \tabitem Explores how \ac{AI} can empower operational decision-making by generating a shared tactical picture, predicting enemy maneuvers, and evaluating own forces options under a limited time. \\ \midrule

2008 & Ref~\cite{surdu2008deep} & \tabitem Describes the Deep Green that concept uses real-time information and simulations to generate flexible, adaptable plans for ongoing military operations, allowing commanders to react to unfolding events. \\ \midrule

2005 & Ref~\cite{bader2005dimensions} & \tabitem Provide a comprehensive overview of the field of Neural-Symbolic integration, including a new classification scheme based on the architecture of the underlying \ac{AI} systems and their capabilities. \\ \midrule

2005 & Ref~\cite{kott2005tools} & \tabitem Describes the use of game-theoretic and deception-resistant algorithms to generate enemy action estimates and the human-in-the-loop wargames used for evaluation. \\ \midrule

\end{tabularx}

\end{table*}

\vspace{1.0ex}

\noindent \textbf{Contributions}. This paper makes the following key contributions to the field of Neuro-Symbolic AI.

\vspace{1.0ex}
\begin{itemize}
    \item Contributing to the growing body of research, this study represents a comprehensive exploration of the extensive possibilities offered by Neuro-Symbolic AI in the context of military applications.

    \item Our paper emphasizes the transformative potential of Neuro-Symbolic AI in military operations, highlighting its role in enhancing national security and operational effectiveness. By providing a comprehensive analysis of its applications and implications, we underscore the significance of integrating Neuro-Symbolic AI into military strategies and decision-making processes.

    \item Our paper comprehensively addresses the ethical, strategic, and technical considerations crucial to the development and deployment of Neuro-Symbolic \ac{AI} in military settings.

    \item Finally, this paper outlines the strategic implications of integrating Neuro-Symbolic \ac{AI} into military operations under dynamic and complex situations, emphasizing how it can redefine strategies, improve operational effectiveness, and contribute to national security.

\end{itemize}

\vspace{1.0ex}

\noindent \textbf{Organization}. The rest of the paper is organized as follows. Section~\ref{neuro_symbolic_AI} explains the theoretical foundations of Neuro-Symbolic \ac{AI}, its practical use cases, and examples of how Neuro-Symbolic \ac{AI} can be applied across various domains. In Section~\ref{autonomy_in_military_weapons_systems} and its subsections, we discuss the role of Neuro-Symbolic AI in enhancing the autonomy of military weapons systems. Section~\ref{military_applications_of_AI} provides a broader perspective on the use of Neuro-Symbolic \ac{AI} in military applications and examples of the implementation of Neuro-Symbolic \ac{AI} in military operations. In Section~\ref{challenges_and_risks} and its subsections, we discuss the identified challenges, risks, and long-term implications associated with the widespread adoption of Neuro-Symbolic \ac{AI} in military contexts. Section~\ref{interpretability_and_explainability} explains the importance of interpretability and explainability of \ac{AI} in military applications. Finally, Section~\ref{conclusion} concludes the paper and outlines our
future directions of research.

\begin{figure*}[h!]
\vskip-0.45cm
	\centering
	\includegraphics[width= \linewidth]{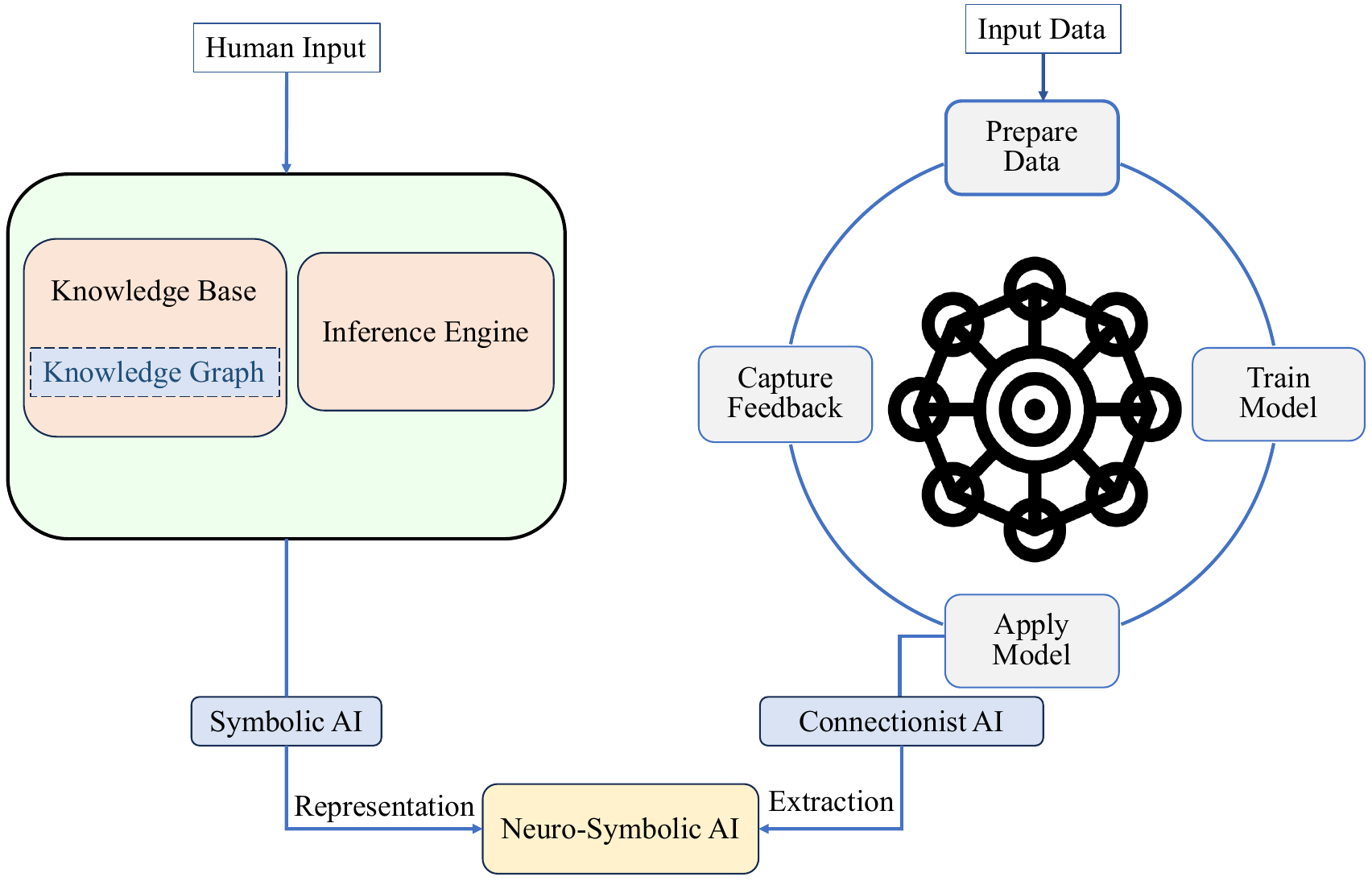}
	\caption{An Example of a Neuro-Symbolic AI Architecture. Note that this is one of many possible architectures in the field.}
	\label{Fig:NeuroSymbolicAI}
\vskip-0.45cm
 
\end{figure*}

\section{Neuro-Symbolic AI}
\label{neuro_symbolic_AI}

Neuro-Symbolic AI's development is influenced by the historical evolution of \ac{AI} paradigms, tracing its roots back to the early days of \ac{AI} research~\cite{goel2022looking}. As previously explained, Neuro-Symbolic \ac{AI} combines neural networks with symbolic reasoning techniques, incorporating both connectionist and symbolic approaches~\cite{hitzler2022neuro}. This fusion allows the system to understand and manipulate symbols while also learning from data. Such systems can be generally represented as shown in Equation~\ref{Eq:NeuroSymbolicAI} where $y_i$ denotes the output of the $i^{th}$ neuron, $f$ the activation function applied to the weighted sum of inputs (as in the purely connectionist model, see Equation~\ref{Eq:feedforward-connectionist-model}), $g$ is a function that combines the neural network output with a symbolic rule, and $z_i$ represents the external symbolic information or rule associated with the $i^{th}$ neuron. The weights $w_{ij}$ determine the influence of input $x_j$ on node $i$, and $b_i$ represents the bias term. In this context, the neural network component ($f$) is responsible for learning from data, while the symbolic component ($z_i$) introduces prior knowledge input or rules. The $g$ function combines both aspects to produce the final output. It is worth noting that the specific form of $g$ can vary widely based on the nature of the symbolic rules and how they are integrated into the underlying neural network. Therefore, it is also worth noting that Equation~\ref{Eq:NeuroSymbolicAI} focuses on a typically single neural network component.

\begin{equation}
y_i=g\left(f\left(\sum_{j=1}^n w_{i j} \cdot x_j+b_i\right), z_i\right)
\label{Eq:NeuroSymbolicAI}
\end{equation}

Neural networks excel at learning complex patterns from data, but they often lack explicit knowledge representation and logical reasoning capabilities~\cite{lecun2015deep, bengio2017deep}. Symbolic reasoning techniques, on the other hand, are well-suited for tasks involving structured knowledge and logic-based reasoning, but they can struggle with data-driven learning and generalization~\cite{russell2010artificial}. While symbolic reasoning handles structured knowledge and logic-based reasoning, it may face challenges when dealing with large and complex problems. In contrast, neural networks efficiently learn from extensive datasets and recognize complex patterns~\cite{anderson1995introduction, bengio2017deep}. Neuro-Symbolic \ac{AI} models typically aim to bridge this gap by integrating neural networks and symbolic reasoning, creating more robust, adaptable, and flexible \ac{AI} systems. In Figure~\ref{Fig:NeuroSymbolicAI}, we present one example of a Neuro-Symbolic AI architecture that integrates symbolic reasoning with neural networks to enhance decision-making. This hybrid approach allows the AI to leverage both the reasoning capabilities of symbolic knowledge and the learning capabilities of neural networks. A key component of this system is a knowledge graph, which acts as a structured network of interconnected concepts and entities. This graph enables the AI to represent relationships between different pieces of information in the knowledge base, facilitating more complex reasoning and inference. As a result, \ac{AI} systems can understand and manipulate symbolic information effectively, reason logically, draw inferences from knowledge bases, learn from extensive datasets, identify complex patterns, generalize knowledge to new situations, and adapt to changing environments. The combination of these two approaches results in a unified knowledge base, with integration occurring at various levels. Examples include incorporating symbolic reasoning modules into neural networks, embedding neural representations into symbolic knowledge graphs, and developing hybrid architectures that seamlessly combine neural and symbolic components~\cite{marino2021krisp}. This enhanced capacity for knowledge representation, reasoning, and learning has the potential to revolutionize \ac{AI} across diverse domains, including natural language understanding~\cite{mcshane2017natural}, robotics, knowledge-based systems, and scientific discovery~\cite{wang2023scientific}. While our paper focuses on a Neuro-Symbolic AI for military applications, it is important to note that the architecture shown in Figure~\ref{Fig:NeuroSymbolicAI} is just one of many possible architectures of a broader and diverse field with many different approaches. Other significant contributions include the architectures described in~\cite{kautz2022third}.

\subsection{Learning Cycle of Neuro-Symbolic AI}

\begin{figure}[h!]
	\centering
	\includegraphics[width= \columnwidth]{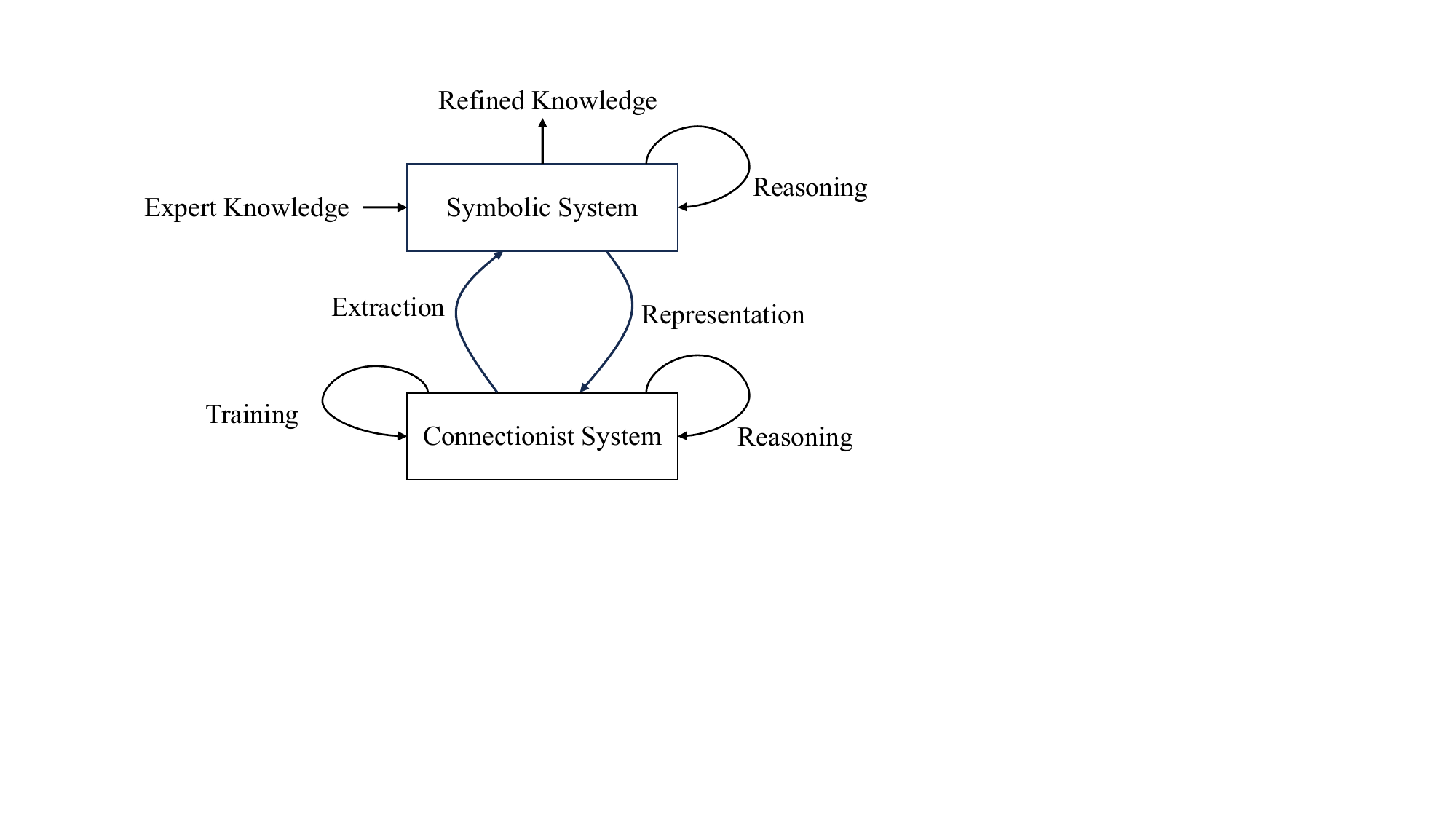}
	\caption{The learning cycle of Neuro-Symbolic \ac{AI} systems~\cite{bader2005dimensions}.}
	\label{Fig:NeuroSymbolicAI_Learning_cycle}
\end{figure}

As shown in Figure~\ref{Fig:NeuroSymbolicAI_Learning_cycle}, the learning cycle of a Neuro-Symbolic \ac{AI} system involves the integration of neural and symbolic components in a coherent and iterative process.

\vspace{1.0ex}

\noindent \textbf{Representation Learning}. In a representation learning setting, neural networks are employed to acquire meaningful representations from raw data. This process often entails training deep neural networks on extensive datasets using advanced \ac{ML} techniques~\cite{bengio2013representation, bengio2017deep}. Representation learning enables networks to automatically extract relevant features and patterns from raw data, effectively transforming it into a more informative representation. This learned representation captures the essential characteristics and features of the data, allowing the network the ability to generalize well to previously unseen examples. Deep neural networks have demonstrated remarkable success in representation learning, particularly in capturing hierarchical and abstract features from diverse datasets~\cite{lecun2015deep, bengio2017deep}. This success has translated into significant contributions across a wide range of tasks, including image classification, \ac{NLP}, and recommender systems. By automatically learning meaningful representations, neural networks can achieve reasonably higher performance on tasks that demand understanding and extraction of relevant information from complex data~\cite{bengio2017deep}.

\vspace{1.0ex}

\noindent \textbf{Symbolic System}. The process of transforming learned neural representations into symbolic representations involves the conversion of neural embeddings into interpretable and logically reasoned symbolic entities~\cite{garcez2008neural}. This transformation is a crucial step in bridging the gap between neural network-based learning and traditional symbolic reasoning~\cite{garcez2022neural}. This approach aims to map neural embeddings, which are distributed numerical representations learned by neural networks, to symbolic entities such as predicates, logical symbols, or rules.  This makes the symbolic representations easier for humans to understand and can be used for tasks that involve logical reasoning~\cite{doran2017does}.

\vspace{1.0ex}

\noindent \textbf{Symbolic Reasoning}. Symbolic reasoning techniques in \ac{AI} involve the use of symbolic representations, such as logic and rules, to model and manipulate knowledge~\cite{galarraga2013amie}. These techniques aim to enable machines to perform logical reasoning and decision-making in a manner that is understandable and explainable to humans~\cite{russell2010artificial}. In symbolic reasoning, information is represented using symbols and their relationships. This encoding approach facilitates the formal expression of knowledge and rules, making it easier to interpret and explain system behavior~\cite{galarraga2013amie}. The symbolic nature of knowledge representation allows human-understandable explanations of reasoning processes. Furthermore, symbolic representations enhance the model transparency, facilitating an understanding of the reasoning behind model decisions. Symbolic knowledge can also be easily shared and integrated with other systems, promoting knowledge transfer and collaboration.

\vspace{1.0ex}

\noindent \textbf{Expert Knowledge}. Incorporating expert knowledge into the underlying AI models can significantly enhance their ability to reason about complex problems and align their outputs with human understanding and expertise~\cite{davis1993knowledge}. This involves domain-specific knowledge, rules, and insights provided by human experts in a particular field. In the symbolic component of a Neuro-Symbolic \ac{AI} system, expert knowledge is often encoded in the form of symbolic rules and logical expressions that capture the structured information and reasoning processes relevant to the application domain~\cite{davis1993knowledge}. Recent advancements have introduced several innovative techniques for incorporating expert knowledge into AI models. \ac{RAG} leverages retrieval mechanisms to enhance the generation capabilities of models by integrating external knowledge sources~\cite{lewis2020retrieval}. G-Retriever employs a novel approach for integrating retrieval-based methods into language models, enhancing their ability to access and utilize domain-specific knowledge~\cite{he2024g}. Additionally, process Knowledge-infused Learning incorporates structured process knowledge into learning algorithms to improve decision-making and reasoning in complex tasks~\cite{sheth2022process}. The effective integration of expert knowledge holds significant promise for addressing complex challenges across various domains, such as healthcare, finance, robotics, and NLP~\cite{garcez2022neural}. For example, expert knowledge plays a crucial role in military operations, enhancing capabilities in strategic planning, tactical decision-making, cybersecurity~\cite{mitra2024localintel, piplai2023knowledge}, logistics, and battlefield medical care~\cite{ben1982knowledge}. Similarly, in a medical diagnosis system, expert knowledge may be encoded as rules describing symptoms and their relationships to specific diseases~\cite{ben1982knowledge}. This enables the AI system to move beyond simple pattern correlation in data and instead engage in reasoning about the underlying medical logic, potentially leading to more accurate and interpretable diagnoses~\cite{ben1982knowledge}.

\vspace{1.0ex}

\noindent \textbf{Refined Knowledge}. In Neuro-Symbolic \ac{AI}, the combination of expert knowledge and the ability to refine that knowledge through iterative learning processes is essential in creating adaptable and effective systems. Expert knowledge serves as a robust initial foundation, while the iterative refinement process allows the model to adapt to new information and continuously enhance its performance~\cite{davis1993knowledge, garcez2002neural}. The iterative process is crucial for enabling the model to adjust to changing conditions, improve accuracy, and address inconsistencies that may arise during the integration of neural and symbolic representations~\cite{garcez2002neural}. It involves continuously updating representations and rules based on feedback from the neural component or real-world data during the training cycle of Neuro-Symbolic \ac{AI}. The continuous learning loop enables the AI to adapt seamlessly to changing environments and incorporate new information. Furthermore, the combined symbolic and neural representation provides insights into the reasoning process and decision-making of the AI, making it more transparent and interpretable for humans~\cite{diaz2022explainable}.

\subsection{Practical Applications of Neuro-Symbolic \ac{AI}}
\label{practical_applications}

Neuro-Symbolic \ac{AI} combines the interpretability and logical reasoning of symbolic 
\ac{AI} with the pattern recognition and learning capabilities of data-driven neural networks, enabling new advancements in various domains~\cite{garcez2023neurosymbolic}. This hybrid approach proves effective in solving complex problems and enhancing decision-making in industrial-strength applications,  including speech and pattern recognition, image analysis, investment analysis, engine monitoring, fault diagnosis, bioinformatics, engineering, and robotics~\cite{bader2005dimensions}. Furthermore, this approach finds practical applications in developing systems that can accurately diagnose diseases, discover drugs, design more efficient NLP networks, and make informed financial decisions. Some of the practical applications of Neuro-Symbolic \ac{AI} include the following.

\vspace{1.0ex}
\noindent \textbf{Natural Language Understanding}. Neuro-Symbolic AI models use a combination of neural networks and symbolic knowledge to enhance the performance of \ac{NLP} tasks such as answering questions~\cite{amizadeh2020neuro}, machine translation~\cite{stahlberg2020neural}, and text summarization. For example, the \ac{NSLM} is a state-of-the-art model that combines a deep learning model with a database of knowledge to answer questions more accurately~\cite{demeter2020just}.

\vspace{1.0ex}
\noindent \textbf{Commonsense Reasoning}. Neuro-Symbolic \ac{AI} can enable \ac{AI} systems to reason about everyday situations, making them better at understanding context~\cite{arabshahi2021conversational, bian2021benchmarking}. This is because Neuro-Symbolic \ac{AI} models combine the strengths of neural networks and symbolic reasoning. One key aspect of commonsense reasoning is counterfactual reasoning, which allows the AI to consider alternative scenarios and their potential outcomes~\cite{barbiero2023interpretable}. This helps the AI understand the cause-and-effect relationships in everyday situations. Another important aspect is defeasible reasoning, where the AI can make conclusions based on the available evidence, acknowledging that these conclusions might be overridden by new information~\cite{furman2023neuro}.

\vspace{1.0ex}
\noindent \textbf{Computer Vision}. Neuro-Symbolic AI models are capable of improving the accuracy and interpretability of various computer vision tasks such as image classification, object detection, scene understanding, and visual reasoning by incorporating symbolic knowledge into neural networks~\cite{yu2022probabilistic, wang2021interpretable, li2021calibrating, amizadeh2020neuro, corysceneunderstanding}. This enables neuro-symbolic AI models to reason about the world and make predictions that are more consistent with human understanding.

\vspace{1.0ex}
\noindent \textbf{Healthcare}. Combining symbolic medical knowledge with neural networks can improve disease diagnosis, drug discovery, and prediction accuracy~\cite{choi2016doctor, drance2021neuro, ashburn2004drug}. This approach has the potential to ultimately make medical AI systems more interpretable, reliable, and generalizable~\cite{rajkomar2018scalable}. For example, the work in~\cite{jiang2020medical} proposes a \ac{RNKN} that combines medical knowledge based on first-order logic for multi-disease diagnosis.

\vspace{1.0ex}
\noindent \textbf{Finance}. Neuro-Symbolic \ac{AI} systems have the potential to revolutionize the financial industry by developing systems that can make better financial decisions~\cite{bouneffouf2022survey}. These systems can help financial institutions in building advanced models for predicting market risks~\cite{hatzilygeroudis2011fuzzy}.

\vspace{1.0ex}
\noindent \textbf{Robotics}. Integrating symbolic reasoning with neural networks can enhance the adaptability and reasoning capabilities of robots~\cite{acharya2023neurosymbolic}. A robot that uses symbolic reasoning can efficiently plan its route through an environment more effectively and adaptively than a robot relying on learning from data~\cite{thrun1991active}. By using its symbolic knowledge of the environment, the robot can determine the best route to reach its destination. Additionally, a robot employing symbolic reasoning better understands and responds to human instructions and feedback~\cite{lemaignan2017artificial}. It uses its symbolic knowledge of human language and behavior to reason about the intended communication.

\vspace{1.0ex}
\noindent \textbf{Military}. Neuro-Symbolic AI can be practically used in various military  situations to make better decisions, analyze intelligence, and control autonomous systems~\cite{voogd2018neuro}. It can provide more interpretable and explainable results for military decision-makers. However, it is important to consider the ethical and legal implications of using AI in the military including concerns related to transparency, accountability, and compliance with international laws and norms.

\section{Autonomy in Military Weapons Systems}
\label{autonomy_in_military_weapons_systems}

Autonomy in military weapons systems refers to the ability of a weapon system, such as vehicles and drones, to operate and make decisions with some degree of independence from human intervention~\cite{taddeo2023comparative}. This involves the use of advanced technologies, often including \ac{AI}, robotics, and \ac{ML}, to enable military weapons to perceive, analyze, plan, and execute actions in a dynamic and complex environment. One of the most significant ways in which \ac{AI} is changing the world in military settings is by enabling the development of autonomous weapons systems~\cite{scharre2018army}. Autonomous weapons systems are weapons that can select and engage targets without human intervention~\cite{guersenzvaig2018autonomous}. While these systems are not yet widely deployed in real-world combat situations, these technologies have the potential to revolutionize warfare and defense. Autonomous weapons systems can be classified into the following two general categories.

\vspace{1.0ex}
\subsection{\ac{LAWS}}

\ac{LAWS} are a class of autonomous weapons systems capable of independently identifying, targeting, and engaging adversaries without direct human control or intervention~\cite{guersenzvaig2018autonomous, asaro2012banning}. These systems rely on a combination of sensor data, \ac{AI} algorithms, and pre-programmed rules to make decisions~\cite{horowitz2015meaningful}. Examples of \ac{LAWS} include autonomous drones~\cite{floreano2015science, kreps2016drones}, cruise missiles~\cite{scharre2015autonomy}, sentry guns~\cite{dresp2023weaponization}, and automated turrets. In the context of \ac{LAWS}, Neuro-Symbolic \ac{AI} involves incorporating neural network components for perception and learning, coupled with symbolic reasoning to handle higher-level cognition and decision-making. 

However, the development of \ac{LAWS} has been a subject of significant ethical, legal, and policy debates due to concerns about the potential for these systems to operate beyond human control~\cite{anderson2017debating}, leading to unintended consequences and violations of international law or the escalation of conflicts outside of direct human control~\cite{righetti2018lethal, egeland2016lethal, guersenzvaig2018autonomous, arkin2008governing}. This poses a serious threat to international stability and the ability of international bodies to manage conflicts~\cite{galliott2020lethal}. Furthermore, the work in~\cite{asaro2012banning} presents the potential human rights implications associated with the use of weapon systems, emphasizing their central role in advocating for a ban on such systems. The author further argues that a requirement inferred from international humanitarian law and human rights law governs armed conflict and implicitly mandates human judgment in lethal decision-making~\cite{asaro2012banning}. The deployment of autonomous weapon systems also poses several critical questions regarding the principles of distinction, proportionality, and military necessity. Additionally, it poses challenges in establishing responsibility and accountability for the use of lethal force, as stipulated in international treaties, implicitly requiring human judgment in lethal decision-making~\cite{asaro2012banning}.

Integrating Neuro-Symbolic \ac{AI} with \ac{LAWS} holds the potential for significant advantages in addressing decision-making complexity and adaptability. However, this integration also amplifies the concerns mentioned above and introduces additional challenges. For a detailed discussion of the ethical, legal, and technical challenges associated with the integration of \ac{AI} and autonomous weapons systems, we refer our readers to Sections~\ref{challenges_and_risks} and~\ref{interpretability_and_explainability}. It is important to recognize that while this integration offers enhanced decision-making capabilities, it raises ethical and legal questions, and poses technical challenges. 

Section~\ref{challenges_and_risks} provides an in-depth exploration of these concerns, addressing issues such as the ethical implications of autonomous decision-making in weapons systems and the legal framework governing their use. Furthermore, Section~\ref{interpretability_and_explainability} emphasizes the importance of interpretability and explainability in Neuro-Symbolic \ac{AI} integrated with \ac{LAWS}. Transparency in decision-making processes is essential to address public concerns and ensure accountability. However, Neuro-symbolic \ac{AI} systems can be complex making it difficult to understand their decision-making processes.  As discussed in Section~\ref{interpretability_and_explainability}, this lack of transparency hinders verification, validation, and accountability, especially for critical decisions in \ac{LAWS}. The integration of Neuro-Symbolic \ac{AI} with \ac{LAWS} presents both opportunities and challenges. A comprehensive understanding of the associated risks and the need for interpretability is crucial for the responsible development and deployment of these systems.

\subsection{\ac{NLAWS}}

\ac{NLAWS} are an evolving class of autonomous weapons systems designed for military and security purposes~\cite{herbert1999non}. The primary goal of these systems is incapacitating or deterring adversaries without causing significant lethality. These systems employ non-lethal means, such as disabling electronics, inducing temporary incapacitation, or utilizing other methods to achieve their objectives. Examples of \ac{NLAWS} include non-lethal weapons such as rubber bullets, tear gas, electromagnetic jammers, \etc Integrating \ac{NLAWS} with Neuro-Symbolic AI presents several challenges, particularly in ensuring the interpretability of decisions for human understanding, accountability, and ethical considerations~\cite{holland2020black, core2006building}. Even though the primary purpose of these systems is non-lethal, their deployment in conflict situations raises significant ethical concerns. \ac{NLAWS} must be able to respond effectively to dynamic and unpredictable scenarios, demanding seamless integration with Neuro-Symbolic AI to facilitate learning and reasoning in complex environments. One emerging approach in this context is reservoir computing, which leverages recurrent neural networks with fixed internal dynamics to process temporal information efficiently. This method enhances the system's ability to handle dynamic inputs and supports the learning and reasoning capabilities required for complex environments~\cite{granato2024bridging}. Robust fail-safes and validation mechanisms are crucial for ensuring safety and reliability, especially when \ac{NLAWS} operates autonomously.

\section{Military Applications of Neuro-Symbolic AI}
\label{military_applications_of_AI}

\begin{figure}[h!]
	\centering
	\includegraphics[width= \columnwidth]{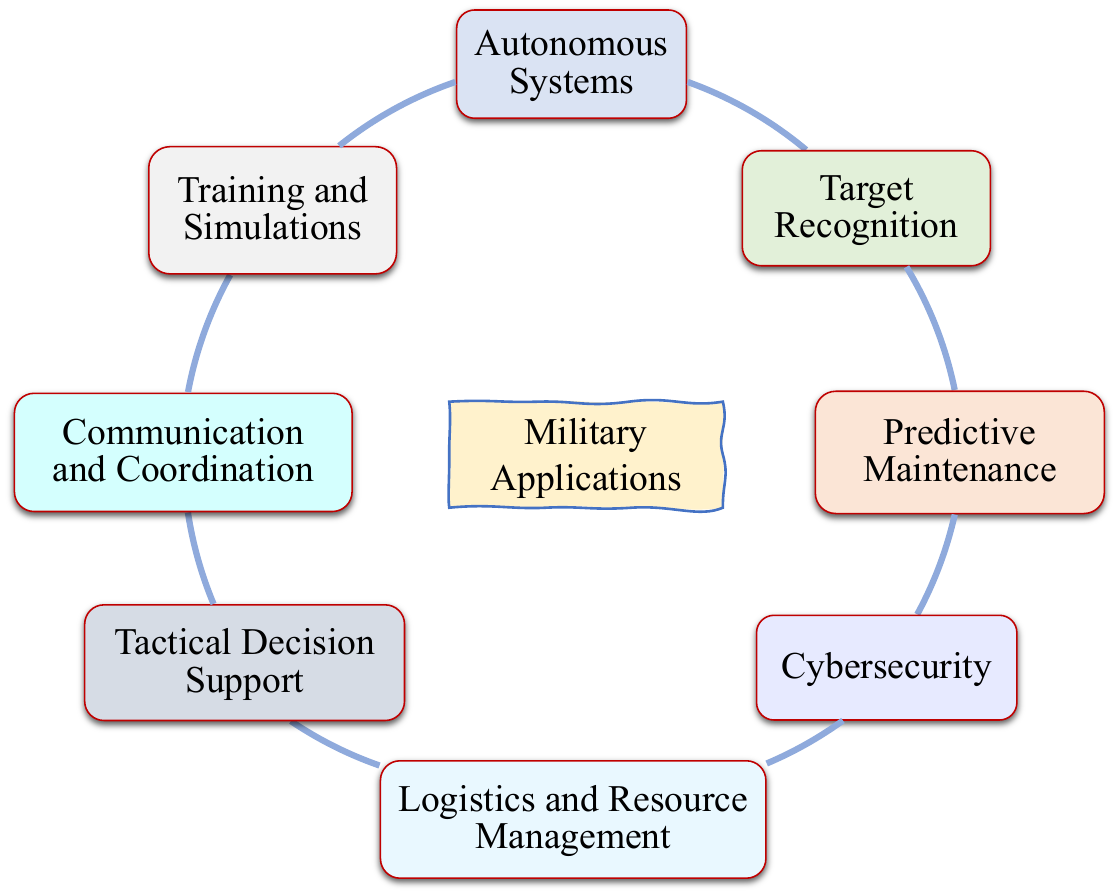}
	\caption{Some of the main military applications of Neuro-Symbolic AI.}
	\label{Fig:Military_Applications_of_NeuroSymbolicAI}
 \vskip-0.45cm
\end{figure}

Militaries worldwide are investing heavily in AI research and development to gain an advantage in future wars. AI has the potential to enhance intelligence collection and accurate analysis, improve cyberwarfare capabilities, and deploy autonomous weapons systems. Figure~\ref{Fig:Military_Applications_of_NeuroSymbolicAI} shows some of the main military applications of Neuro-Symbolic \ac{AI}. These applications offer the potential for increased efficiency, reduced risk, and improved operational effectiveness. However, as discussed in Section~\ref{challenges_and_risks}, they also raise ethical, legal, and security concerns that must be addressed~\cite{righetti2018lethal}.

\subsection{Use Cases}

\noindent \textbf{Autonomous Systems}. Advanced \ac{AI} techniques can be used to develop modern autonomous weapons systems that can operate without human intervention. These AI-powered unmanned vehicles, drones, and robotic systems can execute a wide range of complex tasks, such as reconnaissance, surveillance, and logistics, without human intervention~\cite{arkin2008governing}.

\vspace{1.0ex}
\noindent \textbf{Target Recognition}. One of the key advantages of AI-powered target and object identification systems is that they can automate a task that is traditionally performed by human operators. AI is revolutionizing target and object identification in the military, enabling automated systems to perform this task with unprecedented accuracy and speed~\cite{morgan2020military}.

\vspace{1.0ex}
\noindent \textbf{Predictive Maintenance}. Predictive maintenance is an application of \ac{AI} that leverages data analysis and \ac{ML} techniques to predict when equipment or machinery is likely to fail or require maintenance~\cite{zhang2019data}. \ac{AI} enables predictive maintenance by analyzing data to predict equipment maintenance needs~\cite{keleko2022artificial}. By proactively identifying potential issues in advance, organizations can reduce downtime, minimize unexpected maintenance costs, and optimize their maintenance schedules~\cite{zonta2020predictive}.

\vspace{1.0ex}
\noindent \textbf{Cybersecurity}. \ac{AI} enhances cybersecurity by analyzing patterns, detecting anomalies, and responding rapidly to cyberattacks, thus protecting military networks and information systems~\cite{jajodia2012moving}. Moreover, advanced \ac{AI} techniques help in identifying vulnerabilities in these networks and systems, and to develop and implement security patches and mitigations. By leveraging the capabilities of \ac{AI}, military experts in cybersecurity can contribute to the creation of expert systems that incorporate rules and insights for detecting and responding to cyber threats~\cite{jajodia2012moving}. Experts in military intelligence can provide knowledge about patterns indicative of potential threats. AI systems can then use this knowledge to analyze large datasets, identify unusual patterns, and provide early warnings.

\vspace{1.0ex}
\noindent \textbf{Logistics and Resource Management}. Military logistics experts can provide knowledge about efficient resource allocation and supply chain management.  By leveraging AI-driven systems and advanced strategies, military organizations can use this expertise to optimize logistics, ensuring that resources are deployed effectively during operations~\cite{tsadikovich2010ai, hellingrath2019applications}. Hence, the military can achieve a higher degree of precision in logistics and supply chain management through the integration of \ac{AI} technologies.

\vspace{1.0ex}
\noindent \textbf{Tactical Decision Support}. Through the seamless integration of \ac{AI}, particularly Neuro-Symbolic \ac{AI}, military commanders gain immediate access to real-time data analysis and strategic understanding, enabling more informed and adaptable decision-making on complex battlefields~\cite{cummings2017artificial}. Expert knowledge can be encoded into AI systems to assist military commanders in strategic planning~\cite{geschke1983tac}. This not only improves mission success and reduces collateral damage but also protects soldiers by enhancing potential threat and opportunity identification. By empowering commanders to track troop movements in real-time, analyze communication patterns, and anticipate enemy actions, \ac{AI} contributes to a better understanding of the situation, ultimately leading to superior tactical choices.

\vspace{1.0ex}
\noindent \textbf{Communication and Coordination}. Expert knowledge in military command and control can be used to design advanced AI systems that facilitate effective communication and coordination among different units, enhancing overall operational efficiency. \ac{AI} techniques play a crucial role in improving communication and coordination among military units~\cite{rhodes2005maritime}. By providing real-time data, enhancing situational awareness, and streamlining decision-making processes~\cite{rhodes2005maritime, scott2022enhancing}, these techniques facilitate smoother information flow and faster decision-making during critical moments~\cite{scott2022enhancing}.

\vspace{1.0ex}
\noindent \textbf{Training and Simulations}. Military experts can contribute to the development of realistic training simulations by providing domain-specific knowledge. AI-driven simulations and virtual training environments provide a realistic training experience for military personnel, helping them to develop the skills and knowledge they need to succeed in diverse operational scenarios~\cite{campbell1997use, erickson1985fusing}. This helps in preparing military personnel for various scenarios, improving their decision-making skills, strategic thinking, and ability to handle dynamic and complex situations~\cite{fawkes2017developments}. Beyond training, \ac{AI} can simulate various scenarios, empowering military planners to test strategies and evaluate potential outcomes before actual deployment~\cite{branch2018artificial}.

\subsection{Potential Case Studies}
\noindent \textbf{\ac{ANSR}}. The \ac{DARPA} is funding the \ac{ANSR} research program aimed at developing hybrid AI algorithms that integrate symbolic reasoning with data-driven learning to create robust, assured, and trustworthy systems~\cite{alvaroANSR}. Although the \ac{ANSR} program is still in its early stages, we believe that it has the potential to revolutionize the application of AI use in military operations. ANSR-powered AI systems could be employed to create autonomous systems capable of making complex decisions in uncertain and dynamic environments. For example, ANSR-powered \ac{AI} systems could be used to develop autonomous systems that can make complex decisions in uncertain and dynamic environments. Additionally, ANSR-powered AI systems could be instrumental in developing new tools for intelligence analysis, cyber defense, and mission planning~\cite{alvaroANSR}.

\vspace{1.0ex}
\noindent \textbf{Neuro-Symbolic AI for Operational Decision Support}. The work in~\cite{voogd2018neuro} describes the use of Neuro-Symbolic \ac{AI} in developing a system to support operational decision-making in the context of the \ac{NATO}. The Neuro-Symbolic modeling system, as presented in~\cite{voogd2018neuro}, employs a combination of neural networks and symbolic reasoning to generate and evaluate different courses of action within a simulated battlespace to help commanders make better decisions.

\vspace{1.0ex}
\noindent \textbf{\ac{DG} Concept}. The DARPA's \ac{DG} technology helps commanders discover and evaluate more action alternatives and proactively manage operations~\cite{surdu2008deep, schubert2018artificial}. This concept differs from traditional planning methods in that it creates a new \ac{OODA} loop paradigm. Instead of relying on a priori staff estimates, \ac{DG} maintains a state space graph of possible future states and uses information on the trajectory of the ongoing operation to assess the likelihood of reaching some set of possible future states~\cite{surdu2008deep}. \ac{DG} is based on the idea that commanders need to be able to think ahead and anticipate the possible consequences of their decisions before they are made. This is difficult to do in the complex and fast-paced environment of the modern battlefield. \ac{DG} aims to help military commanders by providing them with tools that can help them facilitate faster decision-making in real-time~\cite{surdu2008deep}. It also helps the commander to identify and assess the risks and benefits of each operation.

\vspace{1.0ex}
\noindent \textbf{\ac{RAID}}. The \ac{RAID} program is another example of Neuro-Symbolic \ac{AI} used in military applications, as discussed in~\cite{kott2005tools}. \ac{RAID}, a \ac{DARPA} research program, focuses on developing AI technology to assist tactical commanders in predicting enemy tactical movements and countering their actions~\cite{kott2005tools}. These include understanding enemy intentions, detecting deception, and providing real-time decision support. \ac{RAID} achieves this by combining \ac{AI} for planning with cognitive modeling, game theory, control theory, and \ac{ML}~\cite{kott2005tools}. These capabilities have significant value in military planning, executing operations, and intelligence analysis.

\section{Challenges and Risks}
\label{challenges_and_risks}

Autonomous weapons systems are considered a promising new technology with the potential to revolutionize warfare~\cite{leys2018autonomous}. However, the development of autonomous weapons systems is raising several ethical and legal concerns~\cite{taddeo2023comparative, anderson2017debating, righetti2018lethal}. For example, there is a concern that \ac{LAWS} could be used to carry out indiscriminate attacks~\cite{taddeo2023comparative}. Furthermore, there is a growing fear that the development of \ac{LAWS} could lead to a new arms race, as countries compete to develop the most advanced autonomous weapons systems~\cite{caton2015autonomous}. Additionally, there are technical challenges to overcome before autonomous weapons systems can be widely deployed~\cite{anderson2013law}, such as reliably distinguishing between combatants and civilians operating in complex environments.

\subsection{Ethical and Legal Challenges}

\noindent \textbf{Moral Concerns}. The integration of \ac{AI} into lethal weapons raises several ethical and moral questions~\cite{lachat1986artificial}, such as discrimination, proportionality, and dehumanization, regarding the moral implications of delegating critical decisions to machines~\cite{pedron2020future, montgomery2017command}. Furthermore, the potential for AI-powered autonomous weapons systems to operate beyond human control poses the risk of unintended consequences, violating ethical norms and principles, such as mission creep~\cite{scherer2015regulating}, hacking and manipulation~\cite{pauwels2020hybrid}, and escalation of conflict~\cite{johnson2020artificial}. One potential approach to address this challenge is to develop comprehensive ethical guidelines and standards for the deployment of Neuro-Symbolic \ac{AI} in military applications. These guidelines should encompass principles of discrimination, proportionality, and accountability~\cite{floridi2021ethical, jobin2019global}. Furthermore, implementing robust monitoring and evaluation mechanisms is crucial to identify and address potential biases or unintended outcomes during AI system deployment~\cite{mitchell2019model}.

\vspace{1.0ex}
\noindent \textbf{Autonomous Decision-Making}. The deployment of Neuro-Symbolic \ac{AI} in military operations raises significant ethical concerns related to autonomous decision-making~\cite{arkin2008governing}. These systems, particularly neural networks, exhibit complex and non-linear behavior that can lead to unforeseen consequences, challenging control, and foreseeability. To mitigate this, ethical considerations must be integrated into the design phase, along with clear guidelines and principles for development and deployment~\cite{floridi2021ethical}. Establishing clear ethical guidelines and principles for developing and deploying Neuro-Symbolic \ac{AI} in military applications can guide responsible decision-making~\cite{jobin2019global}. Integrating ethics into the design phase mitigates potential negative consequences~\cite{floridi2021ethical}. Additionally, implementing robust monitoring and evaluation mechanisms for \ac{AI} systems during deployment is crucial for identifying and addressing potential biases or unintended outcomes~\cite{mitchell2019model}.

\vspace{1.0ex}
\noindent \textbf{Accountability}. The integration of \ac{AI} in military decision-making raises questions about who is ultimately accountable for the actions taken by autonomous systems. It is difficult to hold autonomous weapons systems accountable for their actions under international humanitarian and domestic law~\cite{crootof2016war, beard2013autonomous}. Who is responsible for the actions of an autonomous weapon? For example, who should be held responsible if an autonomous weapons system kills an innocent civilian? Is it the manufacturer, the programmer, or the military commander who ordered the attack? Can a machine be held accountable for killing? The work in~\cite{seixas2022legality} provides a comprehensive analysis of the legality of using autonomous weapons systems under international law. Additionally, it examines the challenges of holding individuals accountable for violations of international humanitarian law involving autonomous weapons systems~\cite{seixas2022legality}. Addressing these challenges requires legal frameworks that clearly define accountability for actions taken by autonomous systems, along with mechanisms to assign responsibility appropriately, whether to manufacturers, programmers, or military commanders~\cite{seixas2022legality, crootof2016war}.

\vspace{1.0ex}
\noindent \textbf{International Humanitarian Law}. Modern autonomous weapon systems raise questions about their impact on various laws, including international human rights law and the right to life. This is particularly evident in policing, crowd control, border security, and military applications~\cite{asaro2012banning}. These systems may also pose challenges in complying with the rules of war, which require the differentiation of combatants from civilians and the avoidance of unnecessary suffering~\cite{anderson2013law, anderson2017debating}. Furthermore, issues related to adherence to principles of distinction, proportionality, and military necessity need to be addressed. Violations of international humanitarian law can result in legal consequences, and ensuring the adherence of Neuro-Symbolic \ac{AI} systems to these principles poses a significant legal challenge in their military use. Developing compliance mechanisms to ensure adherence to international humanitarian law and the principles of distinction, proportionality, and military necessity~\cite{anderson2013law}, along with implementing legal review processes to verify AI system compliance with international laws and norms~\cite{anderson2017debating}, can help mitigate these challenges.

\vspace{1.0ex}
\noindent \textbf{Human Control and Responsibility}. Determining the appropriate level of human control in Neuro-Symbolic \ac{AI}-driven \ac{LAWS} poses challenges~\cite{coeckelbergh2020artificial}. Establishing responsibility and accountability for actions taken by autonomous systems becomes complex, especially in situations requiring human judgment, like navigating ethical dilemmas or exceeding \ac{AI} capabilities~\cite{coeckelbergh2020artificial}. It is, therefore, important to ensure that humans maintain meaningful control over autonomous weapons systems by integrating human-in-the-loop decision-making processes~\cite{boardman2019exploration, holland2020black}. This includes developing override mechanisms to allow humans to intervene and prevent unlawful or unethical actions by autonomous systems~\cite{amoroso2020autonomous}.

\vspace{1.0ex}
\noindent \textbf{Bias and Discrimination}. The training data used for Neuro-Symbolic \ac{AI} models may contain biases, and these biases can be perpetuated in decision-making. This raises ethical concerns related to fairness, equity, and the potential for discriminatory actions, particularly in sensitive military operations~\cite{selbst2019fairness}. Hence, ensuring that Neuro-Symbolic \ac{AI} systems are free from bias potentially leading to discriminatory targeting is essential, especially in complex situations where decisions may impact diverse populations~\cite{ferrer2021bias}. Implementing bias mitigation techniques during the training and deployment of AI models to ensure fairness and equity is crucial~\cite{ferrer2021bias}. Therefore, it is important to use diverse and representative training data to minimize the risk of discriminatory actions by autonomous systems~\cite{ferrer2021bias}. Autonomous weapons systems must be able to reliably distinguish between combatants and civilians, even in complex and unpredictable environments. If autonomous weapons systems cannot make this distinction accurately, they could lead to indiscriminate attacks and civilian casualties violating international humanitarian law~\cite{taddeo2023comparative, anderson2017debating}.

\vspace{1.0ex}
\noindent \textbf{Long-Term Impacts}. The widespread adoption of \ac{AI} in warfare poses a significant challenge given its potential for unforeseen consequences and existential threats in the long-term~\cite{clarke2022survey, brundage2015taking}. Power dynamics among nations may shift dramatically as they leverage \ac{AI}, potentially leading to an arms race and asymmetric conflicts~\cite{bostrom2014superintelligence}. Moreover, unforeseen consequences like loss of control, escalation, and existential threats demand responsible development and international cooperation to mitigate the risks before they become realities~\cite{bostrom2014superintelligence, brundage2015taking}. To address the long-term risks and existential threats posed by AI in warfare ~\cite{bostrom2014superintelligence, brundage2015taking}, fostering international cooperation on developing preventive measures to mitigate loss of control and escalation is crucial.

\vspace{1.0ex}
\noindent \textbf{Global Arms Race}. The development and deployment of Neuro-Symbolic \ac{AI} in the military could lead to an international arms race in \ac{AI}, with nations competing for technological superiority. This race has the potential to intensify geopolitical tensions and reshape global power dynamics. Regulating the rapidly evolving autonomous weapons poses a critical challenge due to the absence of a specific international treaty banning \ac{LAWS} and the difficulty in agreeing on a clear definition~\cite{lewis2014case}. These challenges extend within existing legal frameworks such as the \ac{LOAC} and disarmament agreements designed for human-controlled weapons~\cite{lewis2014case}. The rapid evolution of autonomous weapons creates legal gaps and raises ethical concerns~\cite{taddeo2023comparative}. As nations aim to enhance their capabilities in autonomous weapons systems, there is an increased risk of lowering the threshold for their use, potentially increasing the risk of indiscriminate attacks~\cite{taddeo2023comparative}. Additionally, this development could have diplomatic implications. Clear international regulations and agreements are necessary for governing the use of \ac{AI} technologies in conflict situations~\cite{brundage2020toward, coe2020arms}. To prevent a global arms race in AI-powered weapons, establishing clear international regulations and agreements governing their use in conflicts is crucial~\cite{brundage2020toward, coe2020arms}. Additionally, fostering diplomatic efforts to promote transparency and cooperation among nations regarding developing and deploying autonomous weapons can further mitigate this risk~\cite{taddeo2023comparative}.

\subsection{Technical Challenges}

\vspace{1.0ex}
\noindent \textbf{Knowledge Representation and Integration}. Creating symbolic representations that accurately capture the complexities of real-world battlefield scenarios and their ethical implications is a challenging task~\cite{doare2014robots, fawkes2017developments}. Several factors contribute to this complexity, including the dynamic and unpredictable nature of warfare, uncertainty and incomplete information, and adaptability to changing environments~\cite{boardman2019exploration}. To address these challenges requires developing dynamic models that can dynamically adjust to evolving information and scenarios, effectively manage uncertainty and incompleteness in data, and integrate knowledge from multiple disciplines seamlessly. Furthermore, integrating knowledge from multiple disciplines is crucial for the robustness and accuracy of symbolic representations~\cite{boardman2019exploration}.

\vspace{1.0ex}
\noindent \textbf{Target Discrimination}. In dynamic battlefield environments, accurately identifying combatants and non-combatants is a complex challenge~\cite{skerker2020autonomous}. Ensuring compliance with international humanitarian law and minimizing the risk of civilian casualties are important concerns~\cite{anderson2013law}. Autonomous systems face challenges in low-light conditions, where cameras and advanced sensors may struggle, and radar may misinterpret objects, leading to potential misidentification and harm to civilians~\cite{skerker2020autonomous}. Furthermore, the use of \ac{ML} algorithms trained on biased data introduces the risk of perpetuating discriminatory targeting patterns~\cite{danks2017algorithmic, ferrer2021bias}. For example, an algorithm trained on data identifying combatants with specific ethnicities or clothing styles may erroneously target individuals with similar appearances, regardless of their actual involvement in the conflict. Enhancing target discrimination in diverse conditions can be achieved through advanced sensors and multispectral imaging, coupled with training ML algorithms on unbiased and varied datasets~\cite{danks2017algorithmic, skerker2020autonomous, ferrer2021bias}.

\vspace{1.0ex}
\noindent \textbf{Security and Robustness}. As explained above, nations possessing advanced Neuro-Symbolic \ac{AI} capabilities could gain a strategic advantage. This could lead to concerns about security and potential misuse of \ac{AI} technologies, prompting diplomatic efforts to address these issues. Hence, the security and robustness of autonomous weapons systems are crucial for addressing ethical, legal, and safety concerns~\cite{hall2017autonomous}. Ensuring resistance to cyber threats such as hacking, data manipulation, and spoofing is essential to prevent misuse and unintended consequences~\cite{arkin2008governing, svenmarck2018possibilities}. A reliable, ethical decision-making process, including accurate target identification, proportionality assessment, and adherence to international law, is essential. To enhance the robustness and resilience of Neuro-Symbolic \ac{AI} systems against adversarial attacks, training the underlying \ac{AI} model with both clean and adversarial inputs is effective~\cite{goodfellow2014explaining, madry2017towards}. Additionally, incorporating formal methods for symbolic verification and validation ensures the correctness of symbolic reasoning components~\cite{sun2019formal}. Employing ensemble methods further enhances robustness and makes it challenging for attackers to craft effective adversarial inputs~\cite{tramer2017ensemble}.

\vspace{1.0ex}
\noindent \textbf{Reliability}. The reliability of autonomous weapons systems is crucial in minimizing the risk of unintended consequences~\cite{roff2018trust, scharre2018army}. This involves ensuring the reliability of sensor data, communication systems, and decision-making algorithms. Autonomous weapons systems heavily depend on sensors to perceive their environment. Faulty sensors or misinterpretations can lead to targeting the wrong individuals or objects, leading to civilian casualties, and posing legal and ethical challenges~\cite{buolamwini2018gender}. Sensors and their communication channels are vulnerable to cyberattacks that can manipulate data, causing malfunctions or deliberate targeting of unintended entities. Implementing secure communication protocols and robust cybersecurity measures is essential to safeguard against such manipulations~\cite{scharre2018army}. Furthermore, reliable communication is crucial for transmitting data to and from autonomous weapons systems. The use of redundant communication channels and fail-safe mechanisms is necessary to ensure uninterrupted operation, even in the event of a channel failure~\cite{bhuta2016autonomous}.

\vspace{1.0ex}
\noindent \textbf{Adaptability}. Enhancing the adaptability and robustness of Neuro-Symbolic \ac{AI} systems in unpredictable and adversarial environments is crucial. Warfare is dynamic, and battlefield conditions can evolve rapidly. Warfare is dynamic, and battlefield conditions can evolve rapidly. Therefore, autonomous weapons systems must possess the adaptability to be employed safely in changing and unpredictable environments and scenarios~\cite{anderson2013law}. These systems need to be capable of adjusting their tactics, strategies, and decision-making processes to respond to unforeseen events, tactics, or countermeasures by adversaries. Achieving this level of adaptability requires advanced \ac{AI} algorithms, sensor systems, and the ability to learn from new information and adapt accordingly. One approach to achieving adaptability involves employing advanced \ac{ML} algorithms, such as \ac{RL}, to enable autonomous systems to learn optimal behaviors through trial-and-error interactions with the environment, enhancing their ability to adapt to new and unforeseen situations based on rewards and penalties in dynamic environments~\cite{sutton2018reinforcement}. Transfer learning techniques can also allow Neuro-Symbolic \ac{AI} systems to leverage knowledge from one context and apply it to related contexts, improving their generalization and adaptability capabilities~\cite{pan2009survey}. Additionally, integrating \ac{MAS} can facilitate collaborative decision-making and adaptive behavior in complex environments by enabling multiple autonomous agents to coordinate and share information effectively~\cite{dorri2018multi}.  Continuous monitoring and real-time data integration from diverse sensors can further enhance responsiveness and adaptability by providing up-to-date situational awareness and allowing real-time adjustments to tactics and strategies~\cite{krizhevsky2012imagenet, mnih2015human}. Ensuring explainability and transparency in \ac{AI} decision-making processes remains crucial, especially for autonomous weapons systems. Employing \ac{XAI} techniques can help build trust in the system's adaptation capabilities~\cite{lipton2018mythos}. Additionally, fostering human-AI collaboration, where human operators can intervene and guide the system in complex scenarios, is a promising approach~\cite{leitao2022human, lin2023decision}.

\section{Interpretability and Explainability}
\label{interpretability_and_explainability}

Interpretability and explainability are critical aspects of Neuro-Symbolic \ac{AI} systems, particularly when applied in military settings~\cite{holland2020black, core2006building}. Ensuring interpretability and explainability in advanced Neuro-Symbolic \ac{AI} systems for military applications is important for a wide range of reasons, including accountability, trust, validation, collaboration, and legal compliance~\cite{lipton2018mythos}.

\vspace{1.0ex}
\noindent \textbf{Accountability and Trust}. Transparency and explainability are crucial for algorithms within autonomous weapons systems to build trust and accountability~\cite{balasubramaniam2022transparency}. In the context of military settings, the consequences of \ac{AI} decisions are profound. \ac{XAI} enables military personnel and decision-makers to understand the rationale behind specific \ac{AI} actions, ensuring transparency and building trust in these systems~\cite{holland2020black, core2006building}. This understanding is vital to guarantee alignment with military objectives and adherence to ethical standards~\cite{holland2020black}.

\vspace{1.0ex}
\noindent \textbf{Validation and Error Correction}. Ensuring the reliability, safety, and ethical compliance of \ac{AI} systems is important in military and defense applications. Interpretable \ac{AI} plays a vital role in validating \ac{AI} models and identifying potential errors or biases in their decision-making processes~\cite{holland2020black}, enhancing accuracy, and reducing the risk of unintended outcomes. However, comprehensive testing and verification  remain challenging due to the inherent complexity of military \ac{AI} systems and their potential for unexpected emergent behaviors~\cite{clark2014air}. Recent advancements in Neuro-Symbolic AI have highlighted the importance of robust Verification and Validation (V\&V) methods, Testing and Evaluations (T\&E) processes. Renkhoff et al.~\cite{renkhoff2024survey} provide a comprehensive survey of the state-of-the-art techniques in Neuro-Symbolic T\&E. Their work emphasizes the critical need for rigorous V\&V in these systems, outlining challenges, methodologies, and frameworks to ensure the reliability and safety of Neuro-Symbolic AI applications, particularly relevant for military contexts, in our case, where precision and dependability are important.

\vspace{1.0ex}
\noindent \textbf{Human-Machine Collaboration}. Military decision-making often involves complex tasks that require a combination of human and \ac{AI} capabilities. Interpretable AI facilitates this collaboration between humans and \ac{AI} systems by providing understandable insights into the AI's reasoning~\cite{van2018human, caldwell2022agile}. Such collaboration enhances the overall decision-making process and mission effectiveness, empowering humans to better understand and leverage the AI's insights.

\vspace{1.0ex}
\noindent \textbf{Legal and Ethical Compliance}. Military operations must adhere to international laws and ethical guidelines~\cite{ebers2020regulating, gill2011handbook}. Interpretable \ac{AI}, with its ability to provide transparent insights into AI reasoning, plays a crucial role in ensuring compliance by enabling the military to demonstrate that its decisions align with relevant laws, ethical standards, and the potential consequences of those decisions~\cite{o2019legal}.

\section{Conclusion}
\label{conclusion}

In conclusion, this paper highlights the transformative potential of Neuro-Symbolic \ac{AI} for military applications. This technology could revolutionize modern warfare by enabling complex decision-making in dynamic battlefield conditions, optimizing resource allocation through predictive analytics, and automating intelligence tasks such as target identification and threat assessment. However, the careful development and deployment of Neuro-Symbolic \ac{AI} require  careful consideration of ethical issues, including data privacy, AI decision explainability, and potential unintended consequences of autonomous systems.

Our future work will focus on addressing these challenges while exploring innovative applications such as adaptive robots and resilient autonomous systems. These efforts will advance the role of Neuro-Symbolic \ac{AI} in enhancing national security. We will also investigate optimal human-AI collaboration methods, focusing on human-AI teaming dynamics and designing AI systems that augment human capabilities. This approach ensures that Neuro-Symbolic \ac{AI} serves as a powerful tool to support, rather than replace, human decision-making in military contexts.

\bibliography{NeuroSymbolicAI}\addcontentsline{toc}{section}{\refname}
\bibliographystyle{IEEEtran} 

\end{document}